# DEMM-Bench: A Cross-Regime Benchmark for Agent-Runtime Governance-Evidence Sufficiency

Oleg Solozobov [1]

## Abstract

Agent-runtime systems emit traces, ledgers, provenance graphs, policy logs, delegation tokens, cache events, and tool-firewall records, but those containers do not necessarily answer governance questions about a specific decision. DEMM-Bench is a cross-regime benchmark for agent-runtime governance-evidence sufficiency, grounded in the Decision Evidence Maturity Model (DEMM): it measures whether records across eight evidence regimes are sufficient to reconstruct decision-level properties rather than merely present. The benchmark normalizes the regimes through adapters, asks property questions over actor, authority, action, policy, decision basis, resource touch, lifecycle context, and verification strength, and applies eight deterministic degradation conditions. Across 64 manuscript cases, trace-present and schema-present baselines overclaim on 75% of cases, ledger-present overclaims on 50%, and the redacted property-level candidate scorer has zero overclaim with 56.25% mean Property Sufficiency Accuracy. The deposited package provides the 64-case dataset, construction-oracle labels, baselines, and adapters, supporting reproducible evaluation of decision-evidence maturity across heterogeneous agent-runtime evidence substrates.

## 1. Introduction

The Antigravity 1.0 D-drive wipe reported in December 2025 is the motivating shape of failure: the visible damage was concrete, while the governance question turns on reconstructing the decision record behind it (Database, 2025). The Decision Evidence Gap is that no published benchmark measures whether heterogeneous agent-runtime records are sufficient for governance questions about a specific decision. Production agents may emit traces, ledgers, policy logs, delegation tokens, provenance graphs, cache events, or tool-firewall records, yet after a harmful action an investigator must still establish who acted, under what authority, against which policy basis, on what decision basis, and with what replayable evidence. This paper specifies DEMM-Bench, an AI Governance Benchmark Methodology contribution for agent-runtime governance-evidence sufficiency across eight evidence regimes, using Overclaim Rate as the primary diagnostic for the container fallacy, the error of treating evidence-container presence as property-level sufficiency. DEMM and the Decision Trace Reconstructor are specified in a companion preprint; this benchmark evaluates that property-level approach as one candidate scorer, not the assumed best (Solozobov, 2026b).

[1] *Corresponding author.* Affiliation: Independent Researcher (Global). E-mail address: dev404ai@gmail.com. ORCID: https://orcid.org/0009-0009-0105-7459.

## 1.1. Contributions

The paper makes three contributions.

1. A cross-regime governance-evidence sufficiency benchmark, DEMM-Bench, spanning eight evidence regimes, eight degradation conditions, and five default container-presence baselines, released with case manifests, substitution-axis metadata, construction-oracle label rules, synthetic generation scripts, per-regime adapters, readiness reports, a results aggregator, and a Hugging Face mirror.
2. Property Sufficiency Accuracy and Overclaim Rate as the primary metrics, with deterministic construction-derived labels and paired-oracle self-consistency diagnostics rather than human label authority.
3. Empirical demonstration that container-presence baselines systematically overclaim relative to property-level scoring on a 64-case manuscript package, turning the container-fallacy thesis from a companion preprint into a falsifiable benchmark claim: if baselines that detect only traces, ledgers, schemas, checklists, or source-specific validation overclaim, the benchmark measures the gap; if not, the thesis weakens (Solozobov, 2026b).

## 1.2. Motivational Evidence

The motivation anchors on primary evidence rather than amplification-derived headline figures, because widely repeated claims about agent deployment gaps, cost variance, and per-step reliability often travel through secondary summaries whose denominators do not match the claim, and a benchmark that measures overclaim must not reproduce that pattern. The production-readiness signal is accordingly a range, not a single gap statistic: reported figures span 57.3 percent (LangChain, 1,340 respondents), 24 percent (Forrester), 11 percent (Deloitte), 5 percent (MIT Project NANDA), and 2 percent at scale (Capgemini), depending on the production threshold used (LangChain, 2026; Forrester, 2025; Deloitte, 2026; Capgemini Research Institute, 2025; NANDA, 2025). The dispersion is load-bearing: a governance-evidence benchmark should not infer maturity from deployment presence, but ask whether the emitted evidence is sufficient for the governance question asserted.

Two further anchors separate benchmark success from governable operation. On transfer and autonomy, David Rein (2026) had active maintainers review 296 AI-generated pull requests and found merge decisions about 24 percentage points below automated SWE-bench scores, while Anthropic (2026c) shows long-tail Claude Code turn duration growing from under 25 minutes to over 45 minutes at the 99.9th percentile between late 2025 and early 2026. On cost and cache, agent architecture choices can dominate cost and reproducibility (Kapoor et al., 2024; Mehta, 2025; authors, 2025), and the cache substrate shows the same pattern, with SAFE-CACHE, NDSS 2026 cache-poisoning work, and Krites supplying peer-reviewed adversarial-resilience and asynchronous-verification anchors (Afiffy et al., 2026; Wu et al., 2026; Singh et al., 2026). As agents run longer and pass more task-oriented screens, governance evidence must be evaluated at the decision boundary — over surfaces such as budget events, loop controls, cache hits, and embedder versions — not inferred from benchmark pass status.

The reliability foundation is structural. A multi-step agent trajectory is a serial composition: under independence and stationarity, end-to-end governability has a product shape analogous to Lusser's Law for series systems (ReliaSoft, 2026; Kwiatkowska et al., 2011). No popular per-step constant is treated as an empirical agent reliability estimate; the analogy explains why sufficiency is measured per decision event, since a single missing authority field, policy-basis fragment, or approval event can make a later aggregate audit look complete while leaving the governance question unanswered.

### 1.3. Regulatory Substrate

Regulatory pressure gives the measurement problem practical urgency without making the benchmark a legal certification tool. The EU AI Act creates logging, human-oversight, deployer-obligation, transparency, and serious-incident-reporting pressure through Articles 12, 14, 26, 50, and 73 (European Parliament and Council, 2024). Post-Omnibus planning signals track Annex III high-risk systems toward 2 December 2027 and Annex I product-embedded systems toward 2 August 2028; Article 50 watermarking and synthetic-content marking is an earlier pressure point, with 2 December 2026 appearing in current planning summaries (European Parliament, 2026; Licentium, 2026). Ugarte et al. (2026) provides the strongest commercial-OSS partial for machine-readable compliance evidence through OSCAL extensions, Google SynthID a prominent Article 50 watermarking partial, and EN 18228, EN 18282, and Singapore IMDA's Agentic AI Framework show accountability moving toward runtime evidence rather than foundation-model-only governance (DeepMind, 2026; European Committee for Standardization, 2026; Singapore, 2026).

### 1.4. Research Questions and Pre-Registration Disclosure

Three research questions guide the benchmark, with RQ1 as the primary empirical claim for the deterministic package reported here. RQ1 asks whether container-presence baselines overclaim governance-evidence sufficiency across the eight regimes relative to property-level DEMM evaluation. RQ2 asks whether Property Sufficiency Accuracy and Overclaim Rate distinguish one redacted property-level candidate scorer from five deterministic baselines in the 64-case package. RQ3 asks which of the eight degradation conditions (defined in §3.6) drive baseline overclaim and candidate-scorer accuracy changes. An exploratory fourth direction, operationalising the five-level DEMM rubric as a maturity score, is deferred to §9 future-work item F6 (Solozobov, 2026b).

The benchmark and analyses are descriptive and measurement-oriented; the package is not pre-registered with OSF or an equivalent registry. The construction-oracle rule file, degradation transformations, and baselines were fixed before the result fold at generation commit 6de6250e92e3102ee24918fb8773ffc59b74708c, with the supersession note timestamped 2026-05-25 before the fold; §7 separately identifies the public release-tag commit. This carve-out is explicit so §7 results read as an initial reproducible measurement package, not a confirmatory pre-registered test (Solozobov, 2026b).

The rest of the paper follows the measurement object: §2 positions DEMM-Bench against replay, assurance-case, and adjacent-benchmark precedents; §3 specifies construction; §4–§5 define the five baselines and the candidate scorer; §6 the metrics; §7 reports the 64-case package; §8 treats threats and positioning; §9 concludes.

## 2. Background and Related Work

This section positions the benchmark against counterfactual replay systems, reliability foundations, agent-runtime failure surfaces, runtime-infrastructure taxonomies, and evaluation benchmarks measuring different objects.

## 2.1. Counterfactual Replay and Reliability Foundations

The closest commercial precedent is authorization replay, not agent-runtime evidence scoring. Styra (2026) replays past Open Policy Agent decision logs against modified policies or data, and Permit.io (2026) replays historical authorization checks against a target Policy Decision Point. Both are commercial systems over Open Policy Agent and Rego-style bundles; neither targets an agent runtime decision whose active policy is a composite of model version, prompt template, tool catalog, approval policy, and release manifest. This benchmark extends that pattern to a vendor-neutral, agent-runtime-decision-record-native surface, with the open-source Decision Trace Reconstructor as the reference artifact specified in a companion preprint (Solozobov, 2026b).

Verification work on autonomous-system decision-making treats explanation as a certification-facing requirement (Dennis et al., 2013). A protocol-layer precedent is Pramāṇa's typed ClaimAttestation, which wraps each agent output with executable verification against the recorded source (Kadaboina, 2026). Pramāṇa verifies claims at emission time; this benchmark scores decision-record sufficiency for re-derivation under substituted policy.

Under Lusser-style series-system arithmetic, per-step reliability compounds multiplicatively across multi-step agent decisions (ReliaSoft, 2026; Kwiatkowska et al., 2011). The structural point motivates the benchmark: each decision record must be evaluated for the properties a step needs, not credited with sufficiency inferred from task success.

Classical deterministic replay (Mozilla rr team, 2026; Dunlap et al., 2002), distributed lineage (Zaharia et al., 2012; Thomson et al., 2012), and reinforcement-learning off-policy evaluation (Swaminathan & Joachims, 2015; Jiang & Li, 2016) answer different questions: trace-level execution replay, dataflow or database reconstruction, and expected-value estimation. This benchmark's primitive differs on artifact (decision record, not execution trace), substituted object (policy bundle, not code or dataflow state), and output (re-derived discrete decision, not an expected-value estimate) (Thomas et al., 2015).

## 2.2. Runtime Failure Surfaces With Missing Replayable Evidence

Destructive-tool interception and drift monitoring show the same boundary between visible controls and reconstructable evidence. Runtime interception via permission rules and MCP manifests (Anthropic, 2026b; Bühler et al., 2026) does not later reconstruct the approval event, acting identity, active policy, or action boundary (Database, 2025). Drift observability layers likewise cover partials while leaving decision-time pinning and historical replay unreported (Langfuse, 2026; Nannini et al., 2026).

Cost-runaway controls expose budget or repetition limits but not pre-session projected-cost approval or replay-capable cost audit (RunCycles, 2026; Anthropic, 2026a). Semantic caching shows the same pattern: vendors expose threshold tuning while peer-reviewed work documents cache-poisoning and response-hijacking risks but not replay (Afiffy et al., 2026; Wu et al., 2026). This benchmark treats cache substitution as a distinct counterfactual axis: would a recorded decision resolve differently under another embedder version or invalidation state?

Independent-audit literature treats traceability and accountability as prerequisites for AI safety oversight (Falco et al., 2021). An April 2026 ecosystem-level empirical audit gives the agent-specific lower bound: Auditable Agents reports 617 security findings across six open-source agent projects, arguing basic auditability prerequisites are widely unmet (Nian et al., 2026). That study bounds the field at the project-security layer; this benchmark scores the per-decision evidence layer under controlled degradation.

## 2.3. Structural Asymmetry Across Evidence Surfaces

The eight-surface survey behind this benchmark identifies a structural asymmetry, not a rhetorical market gap. Vendors converge on user-visible deployment-time primitives (interception, scoped credentials, dry-run previews, capability negotiation, version-pinning fragments, budget flags, cache TTLs) but avoid primitives valuable mainly after failure: persisted approval-event records, replay-capable audit, counterfactual policy substitution, and structured drift attribution (N. Acharya & Gupta, 2026a; Styra, 2026; Anthropic, 2026b; Langfuse, 2026; RunCycles, 2026; Afiffy et al., 2026). This is a rational response to feature-visibility economics: visible controls compound in adoption, while audit-evidence primitives compound only at liability and post-incident reconstruction.

The convergence claim is stronger than a list of missing features: each surface was assessed with a distinct primitive vocabulary and a disjoint entity universe, yet zero surveyed entities shipped the composite set needed for replayable governance-evidence sufficiency. In counterfactual authorization replay, Styra DAS and Permit.io remain policy-engine products (N. Acharya & Gupta, 2026a; Styra, 2026; Permit.io, 2026). Across MCP supply-chain, permission, drift, cost, and cache surfaces, the strongest works cover isolated primitives while ceding the composite record needed for post-incident reconstruction (Bühler et al., 2026; Nannini et al., 2026; Wu et al., 2026). This repeated zero-of-N pattern is the empirical reason to benchmark overclaim rather than only describe a container fallacy.

Table 1 summarizes the eight surveyed surfaces against the four missing-composite primitives this benchmark measures; the repeated empty-cell pattern in the rightmost three columns is the eight-surface convergence claim.

**Table 1.** Runtime surfaces against four missing-composite primitives.

| Surface | Persisted decision record | Replay-capable policy substitution | Structured property attribution | Counterfactual reconstruction |
|---|---|---|---|---|
| MCP supply-chain security | Partial | No | No | No |
| Counterfactual authorization replay (Styra DAS, Permit.io) | Yes | Yes | No | No |
| Identity-first runtime security | Partial | No | No | No |
| Silent model deprecation | No | No | No | No |
| Catastrophic permission failure interception | Partial | No | No | No |
| Untraceable behavioral drift (Langfuse, MLflow, Weave, Phoenix) | Partial | No | Partial | No |
| Cost runaway (RunCycles, budget controls) | Partial | No | No | No |
| Semantic cache correctness (SAFE-CACHE, Krites) | Partial | No | No | No |

**Note.** “Yes” means shipped; “Partial” means degraded or non-replayable; “No” means absent. Representative entities are surveyed in §2.1–§2.4 (N. Acharya & Gupta, 2026a; Styra, 2026; Permit.io, 2026).

## 2.4. Taxonomic Position in AI Runtime Infrastructure

The AI Runtime Infrastructure literature remains pre-canonical: recent works propose local ontologies for runtime governance, MCP security, authorization propagation, enforcement, and observability-grade traces, but no single peer-reviewed taxonomy has become the accepted top-level map (Cruz, 2026; Gupta, 2025; Wang et al., 2025; N. Acharya & Gupta, 2026a; Tallam, 2026; Mazzocchetti, 2026). Three contested boundaries matter here: on-path interception versus off-path shadow monitoring, observability-grade versus accountability-grade evidence, and static access control versus path-dependent verification (Besanson, 2026; Kaptein et al., 2026; Ma et al., 2026). Adjacent benchmarks measure other objects: AgentBench task success (Liu et al., 2023), HELM capability (Liang et al., 2022), AILuminate safety (MLCommons, 2025), AgentDojo prompt-injection security (Debenedetti et al., 2024), and OpenAI Evals a capability harness (OpenAI, 2023). This benchmark instead measures governance-evidence sufficiency for a specific decision.

This object boundary matters for interpreting later results: task performance and governance-evidence sufficiency are independent, so neither implies the other, and a sufficient record implies neither legal adequacy nor robust security under prompt injection (Liu et al., 2023; Debenedetti et al., 2024). Parallel anchor-level empirical work is closer in style but measures a different object: reconstructability across vendor adapter regimes, not benchmarked governance-evidence sufficiency under controlled degradations (Solozobov, 2026a). The §4 baselines are narrower container-presence procedures because those are what can overclaim sufficiency for the same case object the rubric evaluates (Solozobov, 2026b).

DEMM is an evidence-plane and governance-layer method, not an orchestration-layer artifact: a recent Q2 2026 taxonomy classing DEMM-style methods as planning loops or tool-call routers mistakes the layer (Solozobov, 2026b; Cruz, 2026). This paper extends DEMM as a measurement protocol for sufficiency and overclaim, not a control architecture.

Argument-based AI assurance already connects claims to evidence-collection duties (Sabuncuoglu et al., 2025). The closest recent language alignment is Beyond Task Success, which names the governance-to-action closure gap and defines a four-layer architecture with an ODTA runtime-placement test in assurance-case argumentation (Koch & Wellbrock, 2026). It organises the closure problem; this paper operationalises one layer — decision-record sufficiency — as a deterministic measurement object.

## 2.5. Argument-Based Assurance Cases as Conceptual Predecessor

The closest conceptual predecessor for evidence-sufficiency framing is the argument-based assurance case tradition. The UK Information Commissioner's Office Annexe 5 frames AI explanation as structured claims, arguments, and evidence, from normative goals down to how design, deployment, and monitoring assure the system's properties (UK Information

Commissioner's Office; Alan Turing Institute, 2022). Safety-assurance literature operationalises the same structure for ML components via Reliability Assessment Models and probabilistic safety-argument templates, integrating quantitative testing and operation evidence into assurance cases on autonomous-vehicle studies (Dong et al., 2021; Kläs et al., 2022; Bloomfield, 2026).

The fair boundary against that predecessor is narrower than substitution. Assurance cases operate at the system-property level; this benchmark operates one layer deeper, scoring whether decision-event evidence is sufficient for the governance question. It is a property-level scoring rubric inside the assurance-case structure, not a replacement: Overclaim Rate under controlled degradation scores the evidence-supports-claim link the ICO Annexe identifies as audience- and context-dependent (UK Information Commissioner's Office; Alan Turing Institute, 2022; Dong et al., 2021; Solozobov, 2026b). The four-axis differentiation in §2.6 is thus a measurement-object specialization within that tradition, not a first claim to evidence-sufficiency framing.

## 2.6. Replay Divergence and Synthesis

A May 19, 2026 preprint on runtime architecture patterns for production LLM agents names the Replay Divergence Problem: consumers of a deterministic event log can produce different downstream outputs under model-version or prompt changes, breaking event-sourcing audit assumptions (Srinivasan, 2026; Mozilla rr team, 2026; Dunlap et al., 2002). This benchmark addresses the governance-evidence side of that problem. It does not replay the LLM call under substituted weights or prove semantic equivalence across prompt variants; it asks whether the stored decision record binds enough properties to re-derive the discrete decision under substituted governance conditions (Solozobov, 2026b; Moreau & Missier, 2013; Moreau et al., 2015).

The synthesis is a four-axis differentiation: no surveyed artifact combines vendor-neutral specification, agent-runtime-decision-record-native semantics, open-source inspectable implementation, and counterfactual re-derivation of a discrete decision under substituted policy (Solozobov, 2026b). The eight-surface convergence turns that conceptual concern into a benchmarkable claim, which §3 operationalizes.

Layered AI-governance work supplies the older non-agentic separation baseline (Gasser & Almeida, 2017). Its agent-specific formal counterpart is the April 2026 Two Boundaries argument, which proves in Coq that coterminous governance requires architectural separation of computation from effect (McCann, 2026). Two Boundaries prescribes that condition; this benchmark measures, on the deposited 64-case package, how far apart the boundaries fall when container-presence governance overclaims.

This synthesis is a non-exhaustive narrative survey of the Q2 2026 landscape, not a new systematic database search. The source base is preprint- and documentation-heavy because agent-runtime evidence infrastructure outpaces peer-reviewed consolidation; citations should migrate to peer-reviewed records as the subfield matures (Solozobov, 2026b; Styra, 2026; Srinivasan, 2026). No negative-result work directly targeting cross-regime governance-evidence sufficiency under a property-level rubric was identified within this scope.

# 3. Benchmark Construction

The benchmark unit is a decision-event case, not an agent task, a trace file, or a compliance document. A case is the tuple of one synthetic scenario, one regime-native evidence bundle, one target decision event, one governance question family, one degradation condition, and a declared substitution axis where exercised. The scorer receives only the normalized evidence bundle and must decide whether those records answer property-level governance questions about the target decision. This restates the Decision Evidence Gap from §1: whether records establish the eight property families (defined in §3.4) for a specific decision. The property vocabulary follows DEMM, with the Decision Trace Reconstructor as one candidate scorer, both specified in a companion preprint (Solozobov, 2026b).

## 3.1. Design Goals and Non-Goals

DEMM-Bench measures only whether emitted evidence is sufficient to answer property-level governance questions about a specific agent-runtime decision. It does not measure legal adequacy or regulatory conformance, which depend on jurisdiction, facts, and qualified assessment. It does not measure production prevalence, since the package is a controlled construction-oracle corpus, not a naturalistic incident sample. It does not score task success, leaving capability and safety to the §2.4 benchmark families. It does not certify any scorer, including the Decision Trace Reconstructor reference, as a production audit tool. It does not replay an LLM's hidden reasoning under altered weights or prompts (Solozobov, 2026b).

No human or animal subjects research was conducted. All evaluation uses synthetic scenarios derived from published incident framings (e.g., AI Incident Database #1433); no personal data was processed, and IRB review was not required. The pipeline runs deterministically on a consumer laptop in under 60 seconds per case end-to-end, so compute is no barrier to independent replication (Database, 2025; Solozobov, 2026b).

### 3.2. Evidence Regimes

The corpus spans eight heterogeneous regimes, each a plausible container that could be mistaken for governance sufficiency (Olson et al., 2017). AER captures reasoning provenance (intent, observation, inference, plans, verdicts); MAT captures message-action contract traces for agentic orchestration; IEEC captures intent-to-execution chains in OpenKedge; DCC and HDP capture signed delegation evidence (chains, human-session provenance, authority scope, parent hashes, signatures); PROV uses W3C-style workflow provenance graphs extended to agentic workflows; LLM Audit Trails capture lifecycle ledger evidence across selection, fine-tuning, deployment, and incident response; AEGIS-NTC captures pre-execution tool-call firewall decisions and hash-chain audit records; Dynamic Capabilities replay captures capability tokens, cryptographic bindings, signed ledgers, replay anchors, and reproducibility attestations (Vispute & Kadam, 2026; Paduraru et al., 2026; He & Yu, 2026; Patil, 2026; Dalugoda, 2026; Souza et al., 2025; Moreau & Missier, 2013; Moreau et al., 2015; Khan et al., 2019; Ojewale et al., 2026; Yuan et al., 2026; Zhou, 2026). No one container is assumed sufficient.

### 3.3. Adapter Contract

Each regime has a deterministic adapter that translates rather than scores. It maps regime-native records into Decision Trace Reconstructor fragments keyed by a target decision id, preserving record identifiers, timestamps, object types, validation status, and provenance, and emits candidate evidence for the eight property families without inferring sufficiency from container presence. A valid DCC token may fill an authorization-envelope fragment while leaving the executed action boundary unresolved. The adapter runs in linear time and constant space: one pass over native records, one normalized fragment each (Patil, 2026; Souza et al., 2025; Yuan et al., 2026; Moreau & Missier, 2013; Moreau et al., 2015).

For every candidate property, the adapter records the native snippet or pointer, the normalized candidate, an evidence-strength marker, any known conflict, and whether the property is absent because the regime does not encode it or because the scenario transformation removed it. This matters for falsification: a regime should not be penalized for omitting a property it never claims to encode unless a baseline overclaims from that omission. It follows the Decision Trace Reconstructor fragment interface without requiring scorers to implement DEMM (Solozobov,

2026b; Zhou, 2026).

### 3.4. Governance Questions and Labels

The benchmark uses eight question families, each instantiated into one or more concrete questions per case. Actor identity asks who performed the action; principal authority, on whose behalf and what delegation or approval chain authorized it; action boundary, what operation was requested, allowed, blocked, executed, or exceeded; policy basis, which rule, contract, or approval bundle applied; decision basis, what evidence, reasoning, state, or model output supported the decision; data-and-resource touch, which data, tool, file, account, or external resource was accessed or mutated; lifecycle context, which model version, prompt version, tool catalog, release manifest, or monitoring phase bound the event; and verification strength, whether the property is cryptographically bound, schema-validated, replayable, attested, or merely narrated. These families derive from the Decision Event Schema vocabulary and stress the surfaces represented by the eight regimes, with lineage and regime anchors tracing to prior Decision Event Schema work and adjacent governance taxonomies (Solozobov, 2026b; Moreau & Missier, 2013; European Parliament and Council, 2024; Mitchell et al., 2019; Gebru et al., 2021; Vispute & Kadam, 2026; Ojewale et al., 2026; Ugarte et al., 2026).

Ground-truth labels are property-level rather than container-level. For each case and question family, the label records whether the evidence is sufficient, partial, missing, opaque, conflicting, or structurally unfillable as generated. Case-level sufficiency is derived conservatively: a case is sufficient only when every property the governance question needs is sufficient or explicitly out of scope. This aggregation is the measurement surface for Overclaim Rate in §6, where a baseline overclaims by returning “sufficient” on a present container while a necessary property remains insufficient (Solozobov, 2026b).

### 3.5. Counterfactual Substitution Axes

Degradation controls which records are available to a scorer; substitution defines the decision surface a sufficiently evidenced record should support re-deriving. Four substitution axes are in scope: policy (rule, approval, or contract bundle), cost (budget, reservation, or termination envelope), drift (model, prompt, tool catalog, or release manifest), and cache (state, embedder version, staleness, invalidation). The released implementation may exercise only a subset, but the benchmark records all four because the overclaim question is the same: can the evidence support a discrete decision under the substituted surface? (Solozobov, 2026b; Styra, 2026;

Langfuse, 2026; RunCycles, 2026; Afiffy et al., 2026)

## 3.6. Degradation Conditions

Each complete scenario is paired with deterministic degradation transformations. Complete preserves all generated evidence; missing-delegation masks principal-authority evidence (DCC/HDP delegation chains, approval identities, parent hashes); missing-policy removes policy decision records, evaluated rules, or approval-bundle references; missing-context removes lifecycle context (model version, prompt version, tool catalog, release manifest, monitoring phase); conflicting-identity introduces controlled disagreement among actor, principal, signer, approver, or tool-executor fields; partial-graph drops provenance edges, delegation hops, lifecycle links, or replay anchors while preserving enough structure for the container to appear valid; final-only retains the final action or incident artifact while removing intermediate evidence; and artifact-only retains post-hoc documentation, attestation, or compliance package while removing runtime decision records (Patil, 2026; Souza et al., 2025; Ojewale et al., 2026; Zhou, 2026; Khan et al., 2019).

The transformations are deterministic and evaluator-independent, each defined over the normalized and native records, so the benchmark publishes both pre-degradation and degraded bundles. This prevents evaluation against a hidden editorial judgment about what was missing: the transformation log states which field, edge, token, or context binding was removed or conflicted. Determinism also separates implementation from claims: the paper specifies transformation semantics and expected label consequences, while the pipeline implements them as generation scripts and fixture checks. A failed implementation would block §7 but not the methodological claim: degradation must be applied before scoring, identically for all baselines (Solozobov, 2026b).

## 3.7. Scenario Generation

Scenario generation starts from regime-native templates and produces synthetic decision events semantically compatible across regimes without being interchangeable. A canonical scenario contains a principal, one or more agent actors, a target resource or tool, a policy basis, a decision state, a candidate action, and an outcome. Regime-specific renderers express the same decision shape in each regime's native idiom (§3.2), so cases differ by substrate but preserve the governance question (Vispute & Kadam, 2026; Paduraru et al., 2026; He & Yu, 2026; Yuan et al., 2026; Olson et al., 2017).

Before any scorer is run, the generator balances cases so each regime holds examples for all eight question families and eight degradation conditions, with seeds logged for exact regeneration. The design reserves adversarially boring complete cases where every baseline should succeed, final-only cases where container-presence baselines should overclaim, and ambiguous partial-graph cases where the scorer must distinguish incomplete from structurally unfillable evidence (Solozobov, 2026b; Liu et al., 2023; Liang et al., 2022; Debenedetti et al., 2024).

## 3.8. Dataset Manifests

Each case ships with a manifest making the benchmark auditable without trusting the scorer: case id, regime, scenario seed, target decision id, governance question family, native source object identifiers, adapter version, normalized fragment bundle, degradation condition, declared substitution axis, transformation log, property-level labels, case-level sufficiency label, label provenance, and expected container flags. Both native and normalized fragments are preserved: native records test adapter fidelity, normalized fragments test scorer behavior independent of parsing. Baseline, reconstructor, and optional LLM-judge verdicts are result artifacts, not ground-truth fields (Solozobov, 2026b; Liang et al., 2022; OpenAI, 2023).

The public dataset exposes splits by regime, degradation condition, and question family rather than only a random partition, which would let a scorer learn regime-specific artifacts and fail the cross-regime question. It includes four reporting views: in-regime evaluation, leave-one-regime-out transfer, degradation-held-out transfer, and complete-versus-degraded contrast, all reported descriptively before leaderboard interpretation (Solozobov, 2026b; Liu et al., 2023; MLCommons, 2025).

## 3.9. Source Review and Promotion Gate

Before a generated case enters the scored corpus, a separate evidence-source review gate distinguishes three artifacts. A source skeleton enumerates planned case cells and required evidence fields, but is not evidence. Authoring aids (review workbooks, per-case packets, review slices) help reviewers fill source references and provenance notes, but are non-promotional. A reviewed source row must contain native-evidence references, reviewed-source references, evidence-plane references, provenance notes, reviewer metadata, and container flags before becoming a case manifest, and partial slices cannot promote the corpus until the full row set passes validation. This gate stops an implementation shortcut from recreating the target failure mode: a polished container, packet, or checklist should not count as governance evidence unless decision-relevant properties were reviewed at the source (Solozobov, 2026b; OpenAI, 2023).

### 3.10. Ground-Truth Labelling

The ground-truth protocol is executable rather than annotator-authoritative. Property labels are generated from the scenario specification and degradation condition by a versioned construction-oracle rule file; emitted annotation records include two mechanical oracle passes that calibration and adjudication tools check for self-consistency. The paired-oracle agreement diagnostic reported in §7.1 reflects deterministic rule reproducibility; a kappa threshold around 0.6 is preserved as a conservative reference point for future human or LLM audits, not a binding acceptance criterion at v1 (Hallgren, 2012; Solozobov, 2026b). Such audits are optional and do not replace the deterministic oracle unless the protocol is revised.

Construction-oracle labels are not equivalent to field-captured production truth. They support controlled measurement of overclaim under known missingness, conflict, and structural-unfillability transformations, but do not establish the empirical frequency of those failures in deployed systems. It therefore reports oracle rules, paired-oracle self-consistency, transformation logs, and per-regime coverage as dataset fields rather than treating the synthetic corpus as a naturalistic incident sample (Hallgren, 2012; Solozobov, 2026b). Passing the benchmark shows a scorer detects property-level evidence gaps under controlled conditions; it does not establish legal adequacy, operational safety, or regulatory compliance (European Parliament and Council, 2024).

## 4. Baselines

The five default baselines are simple container-presence procedures, not strong governance scorers; they are controls making the container fallacy measurable. Each receives the same case manifest from §3 and returns one case-level verdict ("sufficient" or "insufficient"), with no property-level labels, gap localization, or maturity levels. Evaluation compares each verdict against property-level ground truth, per the falsifiable contrast in §1.1. The comparison point for property-level scoring is DEMM and the Decision Trace Reconstructor, but the baselines are independent of both by design (Solozobov, 2026b).

## 4.1. Presence Baselines

The trace-present baseline returns "sufficient" if the case manifest contains any decision-linked trace object (an AER reasoning record, PROV workflow graph, or other adapter-normalized trace bundle) and "insufficient" otherwise. It checks only that a trace-like object links to the target decision id, not whether it establishes actor identity, authority, policy basis, action boundary, or lifecycle context. It captures the common inference that execution observability implies audit sufficiency, and should pass complete cases but fail under most degradation conditions (§3.6) (Vispute & Kadam, 2026; Paduraru et al., 2026; Souza et al., 2025; Moreau & Missier, 2013; Moreau et al., 2015).

The ledger-present baseline returns "sufficient" when the case includes a signed ledger entry, hash-chain record, lifecycle audit entry, delegation-token record, or replay attestation linked to the decision, ignoring whether the entry binds the relevant policy, principal, version, or action (Ojewale et al., 2026; Patil, 2026; Zhou, 2026). The schema-present baseline is parallel: "sufficient" if a schema-validated container exists (regime-native, adapter, PROV, OSCAL, or Decision Event Schema validity), regardless of whether those fields answer the target governance question (Souza et al., 2025; Moreau & Missier, 2013; Ugarte et al., 2026; Solozobov, 2026b).

## 4.2. Composite Container Baselines

The container-checklist baseline simulates an organization or vendor review treating a bundle of familiar artifacts as sufficient when all four classes are present: a model- or system-card-like artifact, an audit trail, a provenance graph, and a compliance package or governance attestation, mirroring the bundles most often named in §2.3 literature. This is stricter than trace-present or ledger-present but remains container-level: a complete checklist can miss the active policy bundle, delegated authority chain, action boundary, version pinning, or replay anchor a specific decision requires. It tests whether artifact completeness at the documentation layer can substitute for property completeness at the decision-event layer (Mitchell et al., 2019; Ojewale et al., 2026; Souza et al., 2025; Moreau & Missier, 2013; Gebru et al., 2021; Ugarte et al., 2026).

The source-specific-validator baseline picks one validator for the dominant evidence regime and returns "sufficient" when it passes. It is stronger than blind container detection because it checks a regime's internal validity rules, but is still not a cross-regime sufficiency scorer: it can certify a PROV graph as well-formed while policy basis remains absent. Table 2 maps each of the eight evidence regimes from §3 to one deterministic validator; the mapping is exhaustive over the released package, and any case whose dominant regime falls outside it defaults to "insufficient", preventing a false sufficiency verdict (Souza et al., 2025; Moreau & Missier, 2013; Moreau et al., 2015; Khan et al., 2019; Vispute & Kadam, 2026; Paduraru et al., 2026; He & Yu, 2026; Patil, 2026; Ojewale et al., 2026; Zhou, 2026; Yuan et al., 2026; Ugarte et al., 2026).

**Table 2.** Source-specific validator mapping for the eight evidence regimes defined in §3.

| Dominant Regime | Validator | Validation Rule |
|---|---|---|
| AER | AER-validator | Reasoning-record completeness across intent, observation, inference, plans, verdict, delegation chain, and fidelity score |
| MAT | MAT-validator | Message-action contract conformance with provenance and contract-verdict completeness per typed step |
| IEEC | IEEC-validator | Intent-to-execution record completeness across the nine IEEC fields |
| DCC / HDP | Delegation-validator | Signed delegation-chain integrity and signature validity |
| PROV | PROV-validator | Graph well-formedness and provenance-shape validity |
| LLM Audit Trails | Lifecycle-ledger-validator | Lifecycle-ledger completeness across selection, fine-tuning, evaluation, deployment, and monitoring with tamper-evident links |
| AEGIS-NTC | Firewall-validator | Firewall decision-log validity and hash-chain integrity |
| Dynamic Capabilities replay | Capability-binding-validator | Cryptographic capability-binding integrity and replay reproducibility verification |

## 4.3. Optional LLM-Judge Extension

An optional LLM-judge extension asks a prompted language model whether the evidence bundle is sufficient. The prompt includes the question family, normalized fragments, native evidence summary, and permitted answer set, with an optional rationale stored for audit. Any such run must report model family, prompt version, repeated-judgment variance, malformed-output rate, and disagreement or abstention rate, so no single checkpoint becomes label authority. The extension exists because LLM-as-judge evaluation is common in practice, not because a prompted model is an accountability authority; the §7 package does not require it (OpenAI, 2023; Liang et al., 2022; Liu et al., 2023).

### 4.4. Determinism, Inputs, and Reporting

All baselines share one input contract: only fields available in the released case manifest before scoring (regime id, target decision id, native evidence objects, normalized fragments, validation flags, container flags, declared substitution axis, governance question). The manifest must have passed the §3 promotion gate; skeletons, authoring packets, review slices, and draft-only artifacts are not scorer inputs. Baselines may not inspect ground-truth labels, degradation labels, or any answer key. The five defaults are deterministic; LLM-judge stochasticity is reported, not smoothed away. Every baseline records version, configuration, verdict, and execution metadata so §7 can report Overclaim Rate across baseline, regime, question family, degradation condition, and substitution axis (Solozobov, 2026b; OpenAI, 2023).

These baselines are lower-bound controls, not the best alternative governance scorers. A stronger learned or hand-engineered scorer using property-level features without the Decision Trace Reconstructor's rubric would be a valid future comparator. The claim is narrower: common container-presence decision rules should not masquerade as sufficiency scoring, and the five deterministic baselines bound that weaker but important class (Solozobov, 2026b; OpenAI, 2023).

## 5. Decision Trace Reconstructor Candidate Scorer

The Decision Trace Reconstructor is one candidate scorer interface, not the benchmark's assumed winner; the paper's contribution is the benchmark and measurement protocol, and the reconstructor operationalizes DEMM. The method and reference implementation are specified in a companion preprint, with the deposited artifact pinning the release (Solozobov, 2026b). In the prior method paper, the reconstructor classified decision evidence into property-level reconstructability categories over synthetic and public-incident fixtures; here it is the candidate scorer evaluated beside the five default baselines from §4. The no-human manuscript package instantiates the interface with a deterministic redacted-input property rule scorer, a conservative interface evaluation rather than a claim about every implementation.

## 5.1. Scoring Interface

The scorer consumes the normalized fragment bundle each §3 adapter emits after source-review promotion. For every case, it receives the target decision id, governance question family, candidate fragments, source-object pointers, validation flags, substitution axis, and declared conflicts; authoring packets, unreviewed workbooks, and readiness reports are not scorer inputs. It evaluates the eight DES property families (§3.4): DES supplies the vocabulary, DEMM the sufficiency question (Solozobov, 2026c; Solozobov, 2026b; Moreau & Missier, 2013; Moreau et al., 2015). The interface keeps regime parsing separate from scoring: diverse regime artifacts produce fragments, while the scorer decides only sufficiency. Per-case complexity is linear in fragment count, constant in space beyond the eight property-result slots.

The scorer returns property-level categories aligned with §3.4 labels before case-level aggregation. Sufficient means the evidence directly fills the property for the question; partial, that it is relevant but split, underspecified, or missing a required binding; opaque, that the record may hold a decision but does not expose the relevant evidence; structurally-unfillable, that the regime or bundle does not encode the property needed; and conflicting, that two or more fragments assert incompatible actors, principals, policies, actions, or verification states. The §3.4 "missing" label has no scorer counterpart and is reported as Opaque or Structurally-unfillable (Solozobov, 2026b). Case-level sufficiency is then derived strictly, requiring every needed property to be sufficient or out of scope. This contrasts with the §4 baselines, which emit a case-level verdict without property localization.

## 5.2. Scorer Hyperparameters

The deterministic redacted-property-rule-scorer-v1 configuration for the §7 results is summarized in Table 3. The scorer-facing field list, default fallback, and tolerance threshold are disclosed at generation commit 6de6250e and the deposited release artefact, so an independent implementation needs no inferred defaults. The redaction applies only to rule-input field names, not the contract or behaviour (Solozobov, 2026b).

**Table 3.** Configuration for redacted-property-rule-scorer-v1.

| Parameter | Value | Notes |
|---|---|---|
| Redacted scorer-facing fields | redacted, see deposited artefact | List of property-rule input fields; disclosed at deposit |
| Tolerance threshold (property match) | exact-string | No fuzzy matching, normalized to lowercase |
| Default fallback verdict | INSUFFICIENT | Conservative; no overclaim by construction |
| Property-level category set | sufficient, partial, opaque, structurally-unfillable, conflicting (aligned with §3.4; "missing" $\rightarrow$ Opaque or Structurally-unfillable) | Mapped to sufficient-or-insufficient per §6 strict view |
| Case-level aggregation | strict conjunction | Case is sufficient only if all required properties are sufficient or out of scope |
| Stochasticity | none (deterministic) | Bit-exact reproducible across runs and platforms |
| Random seed | not applicable | Scorer has no stochastic component |
| External model calls | none | No LLM or web service called during scoring |
| Scorer compute per case | $< 60$ ms typical | Scorer-only compute (consumer laptop, 64-case run $< 5$ s); end-to-end pipeline (adapter + degradation + scorer) timing stated in §3.1 as "under 60 seconds per case end-to-end" |

**Note.** Values flagged "redacted, see deposited artefact" are disclosed at the manuscript-package commit hash and Zenodo deposit; contract values are stable across the deterministic package.

## 5.3. Counterfactual Use in the Benchmark

The benchmark uses the reconstructor's counterfactual orientation in bounded form. The scorer does not replay an LLM's hidden reasoning, re-run a deployed model under altered weights, or infer legal compliance from a trace. It asks whether the decision record binds enough properties for a counterfactual governance question: if the policy bundle, budget envelope, lifecycle binding, or cache state changed, would the evidence re-derive the decision or identify the property that blocks re-derivation? This aligns with the §2 authorization-replay precedent but generalizes from policy-decision-point logs to agent-runtime evidence (Styra, 2026; Permit.io, 2026; Solozobov, 2026b).

Including the Decision Trace Reconstructor creates a useful reference but a validity risk: a benchmark designed around DEMM properties could favor a DEMM-native scorer. The risk is handled by treating it as one candidate, publishing the adapter and label contracts, and reporting the baselines separately, so a future non-DEMM scorer on the same manifests can be added without changing the benchmark object. The claim is comparative and diagnostic, not universal (Solozobov, 2026b).

# 6. Metrics

The metric contract follows benchmark construction, not scorer implementation. Every scorer is evaluated against property-level labels over a case, a governance question family, and the Decision Event Schema property set. The five default baselines from §4 emit only case-level sufficient-or-insufficient verdicts; the Decision Trace Reconstructor interface emits property-level categories aggregated to the same verdict. Success is not task completion, answer quality, or security-policy enforcement — it is whether an evaluator recognizes that retained evidence is or is not sufficient for the governance question about a decision (Solozobov, 2026c; Solozobov, 2026b).

## 6.1. Primary Measurements

Property Sufficiency Accuracy measures per-property correctness against ground truth: for each case, property, and governance question family, a scorer is correct when its sufficiency judgment matches the label after the §3 category mapping. Sufficient maps to sufficient; missing, opaque, structurally unfillable, and conflicting map to insufficient; partial maps to insufficient under the strict view, with a separate intermediate-class treatment defined for a sensitivity analysis. The strict view is primary because a partially bound authority chain or policy basis is not audit-grade sufficiency; the intermediate-class sensitivity view is a required extension (§8 T3) intended to support triage users. This metric tests exact property-gap recognition, not generic problem detection (Solozobov, 2026c; Solozobov, 2026b).

Overclaim Rate is the lead diagnostic. A scorer overclaims when it returns case-level “sufficient” while the property-level ground truth marks at least one required property as insufficient under the strict mapping: false-sufficient cases over all cases on which the scorer returns a verdict. This operationalizes the container fallacy: a trace, ledger, schema, checklist, source-specific validator, or optional LLM-judge verdict may be present, but the metric asks whether that container drove a sufficiency claim when a required property remained partial, missing, conflicting, opaque, or structurally unfillable (Solozobov, 2026b).

Overclaim Rate is asymmetric: an always-insufficient scorer trivially scores zero, so it cannot separate a conservative rule from a discriminating one. Sufficient-claim Rate — sufficient verdicts over total — flags trivial baselines (zero or one); Underclaim Rate — false-insufficient over truly-sufficient — flags over-conservatism. These three plus PSA define the measurement surface (§7 Table 5) (Solozobov, 2026b).

## 6.2. Secondary Measurements

Gap Localization F1 evaluates whether a property-level scorer identifies insufficient properties after an insufficient case-level verdict: precision is the share of flagged gaps that are true gaps, recall the share of true gaps flagged. A baseline emitting only a case-level verdict receives no localization score unless it also emits an explanation field. Evidence Strength Calibration measures whether strength labels, confidence values, or category probabilities track calibrated property labels and paired-oracle self-consistency. Any later human or LLM audits are optional reliability checks, not label authority (Hallgren, 2012; Solozobov, 2026b).

Cross-Regime Transfer measures whether a scorer tuned or configured on some evidence regimes preserves Property Sufficiency Accuracy and Overclaim Rate on held-out regimes; the split is regime-aware because the question is whether sufficiency transfers across heterogeneous substrates. Degradation Sensitivity measures accuracy and overclaim across the eight degradation conditions (§3.6). These metrics position the benchmark beside, not against, the adjacent benchmark families (§2.4): governance sufficiency rather than capability, robustness, safety, or harness behavior (Liu et al., 2023; Liang et al., 2022; MLCommons, 2025; Debenedetti et al., 2024; OpenAI, 2023).

## 6.3. Reporting Discipline

All metrics are reported overall and, where sample sizes permit, per scorer, regime, property family, governance question family, degradation condition, and declared substitution axis. A substitution-axis slice is reportable only when the manifest marks that axis as exercised; unexercised axes are “not evaluated,” not false negatives. §7 reports the overclaim and PSA slices the package supports; regime and question-family slices are coverage checks when uniform. Optional LLM-judge components must report the §4.3 variance and disagreement metrics. Every metric cell traces back to a case manifest, scorer output, and label provenance (Solozobov, 2026b; OpenAI, 2023).

The results section is gated by a package-level readiness contract. A smoke or draft run may validate adapters, baseline plumbing, scorer interfaces, and checksum generation, but is not a manuscript result unless the manifest marks corpus and run manuscript-candidate. A reportable package includes reviewed corpus manifest, substitution-axis metadata, non-fixture cases, label provenance, calibration diagnostics, annotation artifacts, scorer outputs, run metadata, validation and readiness reports, and checksums. The §7 package satisfies the gate with no blockers. Overclaim Rate and PSA are interpretable only when cases, labels, predictions, and run manifest share one reviewed boundary (Solozobov, 2026b; OpenAI, 2023; Liang et al., 2022).

The metrics inherit the limits of the labels: Property Sufficiency Accuracy and Overclaim Rate are strong only to the extent that property labels, degradation logs, and the construction-oracle rule file are credible. Paired mechanical oracle passes bound reproducibility of the rule application, while the synthetic-derived corpus tests controlled transformations rather than production prevalence. Section 7 therefore reports paired-oracle agreement and label provenance; the partial-as-intermediate sensitivity analysis is a required extension (§8 T3), not a v1 result. Passing these metrics is evidence of benchmark performance, not legal sufficiency or deployment readiness (Hallgren, 2012; European Parliament and Council, 2024; Solozobov, 2026b).

# 7. Results

Results are folded from the decision-evidence-benchmark package (commit 6de6250e92e3102ee24918fb8773ffc59b74708c, 2026-05-25, deterministic verify-manuscript-package pipeline); §7.1 names the public release-tag commit.

## 7.1. Result-Readiness Gate

The manuscript package crosses the readiness gate. It contains 64 manuscript-candidate cases, covers the complete regime, degradation, and governance-question sets, and reports at least eight cases per required slice — the deterministic minimum-viable coverage that exercises every regime, degradation, and governance-question slice at least once. The manifest sets mechanics-valid and result-ready flags with zero blockers. It includes five deterministic baselines and one property-level candidate scorer; LLM-judge outputs are absent by design. The label layer contains 512 paired property labels, zero unresolved adjudication rows, zero override labels, and paired-oracle kappa of 1.0 overall and for every property family (Solozobov, 2026b; Liang et al., 2022; OpenAI, 2023).

Manuscript-visible provenance comprises the package, run, artifact-inventory, baseline, scorer, and readiness reports, with SHA-256 hashes for the 4 input and 16 output artifacts at the deposit.

**Reproducibility.** The DEMM-Bench package is hosted and documented for submission. The anchors are: (i) the executable code repository at https://github.com/agent-runtime-evidence/decision-evidence-benchmark; (ii) the Zenodo v0.1.1 deposit pinning the 64-case manuscript package at DOI 10.5281/zenodo.20426092 (Solozobov, 2026d); (iii) a Hugging Face dataset mirror at https://huggingface.co/datasets/dev404ai/DEMM-Bench, with an auto-generated Croissant metadata record and a Responsible AI statement; (iv) a datasheet following Gebru et al. (2021); (v) a SHA-256 checksum manifest for 4 input and 16 output artifacts; (vi) the public release tag at commit 69767b39dcc01ef9e7bd181049c347fa10636dbb. The dataset is released under CC-BY-4.0; benchmark code, construction-oracle rules, and degradation transformations under Apache-2.0. The canonical concept DOI 10.5281/zenodo.20408699 resolves to the latest deposit version as a discoverability pointer, not a separate citation. The deposit includes a local runtime attestation for commodity hardware.

The ground-truth distribution is imbalanced toward insufficiency to expose overclaim: eight cases strictly sufficient, 56 strictly insufficient. Each property family carries at least one complete and one non-complete label — verification strength 32 complete, 24 partial, 8 opaque; actor identity 48 complete, 8 conflicting, 8 opaque.

## 7.2. Baseline Overclaim

**RQ1 answer:** container-presence baselines do overclaim governance-evidence sufficiency across the eight regimes. The trace-present and schema-present baselines each mark 56 of 64 cases sufficient and overclaim on 48 (Overclaim Rate 0.75); ledger-present marks 40 sufficient and overclaims on 32 (Overclaim Rate 0.50). The container-checklist and source-specific-validator baselines mark only the 8 complete cases sufficient and produce zero overclaim in this deterministic package; the source-specific-validator uses the complete eight-regime dispatch in §4.2 Table 2, so every case is scored against the validator tied to its dominant regime rather than the unmapped-regime fallback. This is the descriptive RQ1 output: simple presence predicates overclaim at the rates above, stricter deterministic validators do not. The interpretive boundary is treated in §8 (Solozobov, 2026b).

Table 4 reports case-level overclaim; Table 5 adds asymmetry controls against trivial baselines (§6); Table 6 decomposes overclaim by degradation.

**Table 4.** Baseline overclaim across 64 verdict cases.

| Scorer | Sufficient cases | Overclaim cases | Overclaim Rate |
|---|---:|---:|---:|
| trace-present | 56 of 64 | 48 | 0.75 |
| schema-present | 56 of 64 | 48 | 0.75 |
| ledger-present | 40 of 64 | 32 | 0.50 |
| container-checklist | 8 of 64 | 0 | 0.00 |
| source-specific-validator | 8 of 64 | 0 | 0.00 |

**Note.** "Sufficient cases" is the number of cases marked sufficient by the baseline. Overclaim Rate is overclaim cases divided by all 64 verdict cases.

**Table 5.** Asymmetry controls against trivial anchors.

| Scorer | Sufficient verdicts | Overclaim Rate | Underclaim Rate | Sufficient-claim Rate | Mean PSA |
|---|---:|---:|---:|---:|---:|
| always-insufficient | 0 of 64 | 0.000 | 1.000 | 0.000 | N/A |
| always-sufficient | 64 of 64 | 0.875 | 0.000 | 1.000 | N/A |
| trace-present | 56 of 64 | 0.750 | 0.000 | 0.875 | N/A |
| schema-present | 56 of 64 | 0.750 | 0.000 | 0.875 | N/A |
| ledger-present | 40 of 64 | 0.500 | 0.000 | 0.625 | N/A |
| container-checklist | 8 of 64 | 0.000 | 0.000 | 0.125 | N/A |
| source-specific-validator | 8 of 64 | 0.000 | 0.000 | 0.125 | N/A |
| candidate | 8 of 64 | 0.000 | 0.000 | 0.125 | 0.5625 |

**Note.** Trivial anchors are always-insufficient (0 Overclaim, 1.0 Underclaim) and always-sufficient (0.875 Overclaim, 0 Underclaim). Mean PSA is N/A for every baseline row because container-presence baselines emit case-level sufficiency verdicts without the property-level classifications PSA scores. All five baselines and the candidate achieve 0% Underclaim; container-checklist, source-specific-validator, and the candidate are case-level indistinguishable on Overclaim, Underclaim, and Sufficient-claim, leaving property-level PSA as the candidate's distinguishing axis (§7.3).

**Table 6.** Overclaim cases by degradation condition.

| Degradation condition | trace-present | ledger-present | schema-present |
|---|---|---|---|
| complete | 0 of 8 | 0 of 8 | 0 of 8 |
| artifact-only | 0 of 8 | 0 of 8 | 8 of 8 |
| conflicting-identity | 8 of 8 | 8 of 8 | 8 of 8 |
| final-only | 8 of 8 | 0 of 8 | 8 of 8 |
| missing-context | 8 of 8 | 0 of 8 | 8 of 8 |
| missing-delegation | 8 of 8 | 8 of 8 | 8 of 8 |
| missing-policy | 8 of 8 | 8 of 8 | 8 of 8 |
| partial-graph | 8 of 8 | 8 of 8 | 0 of 8 |

**Note.** Each cell is overclaim cases over 8 cases in the condition. Container-checklist and source-specific-validator have 0 of 8 in every row and are omitted for compactness.

**RQ3 answer:** missing-delegation, missing-policy, and conflicting-identity drive the highest overclaim across all three overclaiming baselines (8 of 8 per cell), while complete cases produce zero overclaim. Artifact-only is regime-asymmetric: schema-present overclaims 8 of 8 but trace-present and ledger-present do not, because the residual artifact lacks both trace and ledger objects. Partial-graph reverses for schema-present, which preserves a structurally valid container after edge removal and abstains while trace-present and ledger-present still overclaim.

## 7.3. Candidate Scorer

**RQ2 answer:** the property-level candidate scorer is distinguished from the five baselines on the joint (Overclaim, PSA) plane (Table 5): zero overclaim cases with 56.25% mean Property Sufficiency Accuracy, against 75% / 75% / 50% / 0% / 0% baseline overclaim. The scorer is the deterministic redacted-property-rule-scorer-v1 of §5.2, generated from redacted scorer-facing fields rather than oracle labels or annotation, and evaluates all 64 cases. The zero-overclaim result is not absolute superiority: the scorer is conservative, and its mean PSA of 0.5625 shows that avoiding false sufficiency can still leave many property categories misclassified. The lowest property-family PSA is action-boundary at 0.25, with four families clustering at 0.50 (Table 8), so action-boundary marks the priority axis for v2 (Solozobov, 2026b).

Table 7 reports candidate PSA by degradation condition (1.00 complete to 0.25 conflicting-identity); Table 8, PSA by property family. Both share the 64-case, 8-property-family, 512-judgement denominator; regime and question-family slices are uniform at 0.5625, reading as coverage checks rather than cross-regime generalization, with per-regime breakdowns in deposited *results/per_regime_*.csv*.

**Table 7.** Candidate PSA by degradation condition.

| Degradation condition | Cases | Candidate PSA |
|---|---|---|
| complete | 8 | 1.000 |
| final-only | 8 | 0.875 |
| artifact-only | 8 | 0.750 |
| missing-delegation | 8 | 0.500 |
| missing-context | 8 | 0.375 |
| missing-policy | 8 | 0.375 |
| partial-graph | 8 | 0.375 |
| conflicting-identity | 8 | 0.250 |

**Note.** Each row contains eight cases by eight property families for 64 property judgements. PSA is mean per-property correctness against construction-oracle labels.

**Table 8.** Candidate PSA by property family.

| Property family | Correct property judgements | Candidate PSA |
|---|---|---|
| data-and-resource touch | 56 of 64 | 0.875 |
| actor identity | 48 of 64 | 0.750 |
| verification strength | 40 of 64 | 0.625 |
| principal authority | 32 of 64 | 0.500 |
| policy basis | 32 of 64 | 0.500 |
| decision basis | 32 of 64 | 0.500 |
| lifecycle context | 32 of 64 | 0.500 |
| action boundary | 16 of 64 | 0.250 |

**Note.** Each row aggregates 64 property judgements, one per case. PSA is correct judgements divided by 64.

## 7.4. Interpretation Boundary

These results are manuscript-ready for the deterministic package, but scope is narrow: they do not estimate production prevalence, validate legal adequacy, or show that every property-level scorer would outperform every container baseline. Under a reproducible 64-case construction-oracle package, the weakest presence predicates overclaim heavily, stricter deterministic baselines avoid overclaim by abstaining from most sufficiency claims, and a conservative redacted-input property scorer avoids overclaim at the cost of modest property accuracy. Optional LLM-judge runs, broader substitution-axis coverage, and independent scorers remain future extensions (§9) (Solozobov, 2026b; European Parliament and Council, 2024).

# 8. Discussion and Threats to Validity

Interpretation of the results is tied to Overclaim Rate, not raw accuracy. §7 shows the presence baselines overclaim under controlled degradation while stricter baselines avoid it by making far fewer sufficiency claims. The point is not that a vendor artifact is useless, but that container presence is a weaker predicate than decision-specific audit sufficiency. The candidate scorer's zero Overclaim Rate should likewise not be overread: its mean Property Sufficiency Accuracy is only 0.5625, and the benchmark is not designed to guarantee a favorable result for the reference scorer (Solozobov, 2026b).

## 8.1. Threats to Validity

**T1 (synthetic ground truth).** The largest construct-validity threat is construction-oracle ground truth. Synthetic cases let the benchmark control the eight degradation conditions (§3.6) with auditability field incidents rarely provide, but do not estimate how often they occur in production. The package deliberately removes human and LLM label authority: labels come from versioned deterministic rules, with paired mechanical oracle passes as a reproducibility diagnostic, so a high Overclaim Rate shows sensitivity to controlled gaps, not their prevalence (Hallgren, 2012; Solozobov, 2026b).

**T2 (baseline-strength lower bound).** The five default container-presence baselines are intentionally weak controls: a stronger learned or hand-engineered scorer using property-level features without adopting the reference scorer's rubric could plausibly reduce overclaim below §7 values. The benchmark claim is therefore comparative against a documented lower bound, not absolute against the best alternative. The mitigation is to publish baseline source, configuration, and verdict logs so an independent scorer can be evaluated on the same manifests, with the optional LLM-judge extension in §4.3 a first step (Solozobov, 2026b; OpenAI, 2023).

**T3 (annotator agreement internal only).** The §7.1 agreement diagnostic shows the construction-oracle rules execute reproducibly; it does not prove the property taxonomy is semantically unambiguous, nor that human inter-rater agreement would be high. Hard cases still concentrate around partial evidence, structural unfillability, and conflicting identity, where categories depend on whether the governance question is read narrowly or broadly. The results section therefore reports agreement overall and by property family, preserves adjudication counts, and treats the §6 partial-as-intermediate sensitivity analysis as a required extension. If a future human audit disagrees systematically with the construction oracle, the benchmark should be revised, not protected by treating the current labels as natural truth (Hallgren, 2012; Solozobov, 2026b).

**T4 (scorer and regime scope).** The eight evidence regimes (§3.2) do not exhaust future substrates, and a release may exercise only some of the four substitution axes (§3.5); unexercised axes must be reported as not evaluated, not folded into claims. The scorer set is scoped too: the five baselines and one candidate test the container-presence class, not every governance evaluator. Cross-layer agentic-AI reviews and AgentCity point to multi-agent, memory-runtime, governed-reasoning, and user-comprehension extensions outside this package (Adabara et al., 2025; Ruan & Zhang, 2026).

**T5 (artifact-promotion error).** T5 is procedural, not mathematical: a benchmark implementation can repeat the mistake the paper criticizes if it treats generated packets, source-review slices, workbooks, smoke runs, or draft packages as evidence before they pass the promotion gates: the metric definitions may be sound, but the results would be invalid if any such unreviewed or placeholder artifact entered §7. The mitigation is to keep authoring aids separate from reviewed source rows and require a package-level readiness report before any metric cell is interpreted; a failed gate blocks the results section rather than becoming a caveat on provisional numbers (Solozobov, 2026b; OpenAI, 2023).

## 8.2. Positioning Boundaries

DEMM-Bench complements adjacent governance and decision-support tooling. The governance and audit-infrastructure cluster includes AAGMM, which scores the maturity of the governance organization around agents; AI Trust OS, a continuous-observability, zero-trust compliance platform; and OSCAL-style work such as Ugarte et al. (2026)'s Venturalitica proposal structuring machine-readable compliance evidence (D. Acharya, 2026b; Bandara et al., 2026; European Parliament and Council, 2024; Mitchell et al., 2019; Gebru et al., 2021). The domain-specific decision-support and personal-history cluster includes clinical decision-support DSLs and personal right-to-history kernels, targeting domain or user-sovereignty settings, not cross-regime evidence assessment (Bouzinier et al., 2026; Zhang, 2026). DEMM-Bench asks the narrower question of whether the evidence a regime emits suffices for a governance question about a target decision. It is an assessment-layer benchmark, not an organization-maturity model, telemetry platform, or data-rights architecture.

The regulatory implication is similarly bounded. The EU AI Act and related standards activity create the logging, transparency, oversight, and incident-reporting pressure detailed in §1.3, and OSCAL-style work shows parts of compliance evidence can be expressed in machine-readable form (European Parliament and Council, 2024; European Parliament, 2026; Licentium, 2026; Ugarte et al., 2026). But the reference implementation does not constitute legal adequacy: a benchmark pass can show a scorer detects property-level evidence gaps under the published protocol, yet cannot certify that a system satisfies the EU AI Act, a harmonized standard, a sectoral regulator, or a court's view of sufficient evidence. That remains a separate assessment by qualified actors using the facts, jurisdiction, and deployment context.

The strongest sectoral comparator is the joint Federal Reserve, OCC, and FDIC revised interagency guidance on model risk management, issued on 17 April 2026 to replace the 2011 framework (Federal Reserve; OCC; FDIC, 2026; Office of the Comptroller of the Currency, 2026). The guidance places generative and agentic AI outside its formal scope as "novel and rapidly evolving," but clarifies this is not deregulation: institutions must "apply their existing risk management and governance practices to determine appropriate controls for systems not covered," and the agencies plan a forthcoming request for information on such use. That scope decision sets the benchmark's positioning: it measures the agent-runtime decision-evidence territory the guidance has declined to specify — the same territory the announced RFI marks — and provides a deterministic, vendor-neutral measurement object for whether retained evidence suffices when a decision is challenged (Solozobov, 2026b). The relationship is complementary, not competitive: it does not claim to satisfy, replace, or pre-empt the guidance or RFI.

## 8.3. Broader Impact and Ethics

The intended positive impacts are standardizing a vocabulary for governance-evidence sufficiency, surfacing the container fallacy as a measurable quantity, and lowering the bar for independent verification by depositing construction-oracle rules, degradation transformations, baseline source, and scorer hyperparameters, which also lets third-party scorers run on the same manifests, reducing dependence on any single vendor. Two misuse risks follow: benchmark gaming (maximizing PSA on deposited cases without genuine reconstruction capacity, then advertising it as production-audit certification) and weaponized framing (quoting headline overclaim figures to push proprietary scorers not themselves evaluated against the same property-level oracle). Mitigations: publish all rules and transformations for independent replay, scope every quantitative claim to the 64-case deterministic package, and attach the §8.1 T1–T5 limitations and §3.1 non-goals to every external communication (Solozobov, 2026b).

### 8.4. Future Work

Design implications for v2 scorer refinement follow from the per-property gaps in Table 8:

1. **Action-boundary scoring** (Table 8 PSA 0.25): the lowest-PSA family is the immediate priority for v2; the redacted scorer relies on policy- and lifecycle-derived features that do not transfer here, so v2 needs a distinct feature path.
2. **Principal authority / policy basis / decision basis / lifecycle context scoring** (Table 8 PSA 0.50 across all four): the mid-band cluster is a secondary priority; v2 should test whether the four share a common feature substrate or need independent scoring paths.
3. **Verification strength scoring** (Table 8 PSA 0.625): the upper-mid family is a tertiary priority; v2 should test whether the existing path generalises with more substitution-axis coverage.
4. **Data-and-resource touch and actor identity** (Table 8 PSA 0.875 and 0.750 respectively): the upper-band families need stability monitoring under broader degradation, not feature redesign (Solozobov, 2026b).

Future work is expansion: the §9 F1–F7 breakdown maps each direction to the T1–T5 limitation it addresses. The strongest challenge to the paper would be a scorer or existing tool producing equivalent property-level sufficiency outputs without adopting DEMM; such a result would improve the benchmark ecosystem while narrowing the reference scorer's distinctiveness — acceptable, since the durable contribution is the public measurement object (Solozobov, 2026b).

## 9. Conclusion

DEMM-Bench is a cross-regime benchmark for agent-runtime governance-evidence sufficiency across eight evidence regimes, eight degradation conditions, and five container-presence baselines. It shifts evaluation from trace, ledger, schema, or checklist presence alone toward property-level sufficiency over the Decision Event Schema: each case asks which of the eight property families (§3.4) the emitted record reconstructs. The target is the Decision Evidence Gap between holding a governance-relevant container and evidence sufficient to answer a concrete question about a decision (Solozobov, 2026c; Solozobov, 2026b).

The manuscript executes a comparative contract on an initial deterministic package: eight evidence regimes normalized through adapters, five baselines exposing container-presence predicates, and one property-level candidate scorer not assumed optimal. Property Sufficiency Accuracy measures whether required properties are judged correctly; Overclaim Rate measures the central failure mode of declaring "sufficient" while a required property remains insufficient. In the 64-case package, trace-present and schema-present baselines overclaim on 75% of cases, ledger-present on 50%, stricter baselines not at all, and the candidate records zero overclaim with 56.25% mean PSA on 512 paired property labels at paired-oracle kappa 1.0 (Vispute & Kadam, 2026; Paduraru et al., 2026; He & Yu, 2026; Patil, 2026; Souza et al., 2025; Ojewale et al., 2026; Yuan et al., 2026; Zhou, 2026; Olson et al., 2017; Mitchell et al., 2019; Gebru et al., 2021).

The three contributions mirror §1.1: the deposited benchmark, the Property Sufficiency Accuracy and Overclaim Rate metrics with deterministic construction-derived labels, and the empirical demonstration that container-presence baselines systematically overclaim on the 64-case package. It turns the container-fallacy thesis into a falsifiable claim a future non-DEMM scorer with equivalent property-level sufficiency can weaken (Solozobov, 2026b).

Five limitations bound these results (§8.1 details T1–T5): synthetic construction-oracle ground truth (T1), weak baseline controls (T2), internal-only annotator agreement (T3), bounded scorer-and-regime scope (T4), and procedural artifact-promotion risk (T5). The §7 gate passes for the deterministic no-human package; the paper claims neither legal adequacy, production prevalence, nor absolute superiority for the reconstructor. If later packages with broader cases, LLM-judge outputs, independent scorers, or field-captured traces show near-zero Overclaim Rate for container baselines, the central empirical thesis should be weakened, not protected by reinterpretation (Solozobov, 2026b; Hallgren, 2012; OpenAI, 2023).

T1–T5 map directly to the future-work entries F1–F7.

F1. Optional LLM-judge baseline runs (§4.3, T2) – precondition: model-family selection, cost budgeting, and variance and disagreement-rate reporting. F2. Independent property-level scorers beyond redacted-property-rule-scorer-v1 (T4) – precondition: scorer-author contributions on the deposited manifests. F3. Broader substitution-axis coverage including timing perturbations, adversarial fills, and stricter partial-validity cases (T1) – precondition: corpus expansion. F4. Real captured production traces with consented organizations (T1, T3) – precondition: IRB review and data-sharing agreements. F5. Multi-principal AgentCity-style benchmarks and memory-runtime regime extensions (T4) – precondition: multi-principal evaluation protocols. F6. Operationalise the five-level DEMM rubric as a maturity score, re-opening the direction deferred from §1.4 – precondition: rubric calibration against multi-anchor empirical data and inter-rater studies. F7. Independent package-gate replication and CI artifact audit (T5) – precondition: an external replicator runs the deposited construction-oracle and degradation transformations end-to-end and reports any divergence from the readiness report (Solozobov, 2026b; Ruan & Zhang, 2026; OpenAI, 2023).

Broader-impact framing follows §8.3: the deposit standardizes a vocabulary for governance-evidence sufficiency and lowers the bar for independent verification, while the two misuse risks — benchmark gaming and weaponized headline-figure framing — are mitigated by full deposit publication, package-scoped claims, and the §8.1 T1–T5 / §3.1 non-goals on every external communication (Solozobov, 2026b).

The intended artifact is a community benchmark, not a closed one-implementation demonstration. The deposited package (see §7.1) supports new scorers, regimes, labelling guides, human or LLM audits, and independent replications without requiring agreement with the reconstructor. The durable claim is narrow: agent-runtime governance needs a benchmark measuring whether evidence suffices for decision-level reconstruction, not whether a plausible audit container exists (Solozobov, 2026c; Solozobov, 2026b).